\documentclass[sigconf]{acmart}
\renewcommand{\abstractname}{\MakeUppercase{Abstract}} 
\renewcommand{\refname}{REFERENCES} 
\renewcommand\footnotetextcopyrightpermission[1]{} 
\AtBeginDocument{%
  }

\setcopyright{none}
\copyrightyear{2018}
\acmYear{2018}
\acmDOI{XXXXXXX.XXXXXXX}
\acmConference[2nd HEAL Workshop at CHI Conference on Human Factors in Computing Systems]{Make sure to enter the correct
  conference title from your rights confirmation email}{April 26, 2025}{Yokohama, Japan}
\acmISBN{978-1-4503-XXXX-X/2018/06}

\usepackage{cases}
\usepackage{amsmath}
\begin{document}


\title[Gamified Evaluation and Recruitment Platform]{A Gamified Evaluation and Recruitment Platform for Low Resource Language Machine Translation Systems }


\author{Carlos Rafael Catalan}
\email{c.catalan@samsung.com}
\affiliation{
  \institution{Samsung Research Philippines}
  \country{Philippines}
}

\begin{abstract}

Human evaluators provide necessary contributions in evaluating large language models. In the context of Machine Translation (MT) systems for low-resource languages (LRLs), this is made even more apparent since popular automated metrics tend to be string-based, and therefore do not provide a full picture of the nuances of the behavior of the system. Human evaluators, when equipped with the necessary expertise of the language, will be able to test for adequacy, fluency, and other important metrics. However, the low resource nature of the language means that both datasets and evaluators are in short supply. This presents the following conundrum: How can developers of MT systems for these LRLs find adequate human evaluators and datasets? This paper first presents a comprehensive review of existing evaluation procedures, with the objective of producing a design proposal for a platform that addresses the resource gap in terms of datasets and evaluators in developing MT systems. The result is a design for a recruitment and gamified evaluation platform for developers of MT systems. Challenges are also discussed in terms of evaluating this platform, as well as its possible applications in the wider scope of Natural Language Processing (NLP) research.

\end{abstract}


\maketitle

\section{INTRODUCTION}

Machine Translation Systems are tools used to generate translations from a source language to a target language. While there have been multiple approaches in developing these systems, recently more popular approaches have emerged such as Statistical MT (SMT) and Neural MT (NMT). SMT uses a Language and Translation model to analyze the statistical relationships between the source and target text \cite{machinetranslate_SMT} while NMT uses a neural network to create word embeddings that clusters information to disambiguate the word semantics and generate translations \cite{machinetranslate_NMT}. Since both methods require the use of large quantities of parallel multilingual training data, developing MT systems prove to be difficult when data is scarce, such is the case for LRLs.

Evaluating MT systems can be done through automated metrics and human evaluation. Using automated metrics can be a cost effective way to evaluate MT systems as the quality can be computed quickly. However scores can be unreliable, unexplainable, and may not correlate well with human evaluation scores. \cite{machinetranslate_metrics}. Human evaluation on the other hand is seen as the gold standard of evaluation \cite{machinetranslate_metrics} as a native speaker equipped with the familiarity of the language would be able to adequately evaluate the generated output in terms of both syntax and semantics. In LRLs, like datasets, native speakers are scarce which can pose challenges in conducting an adequate human evaluation.

The author of this paper presents the following research question: How can we increase the representation of human evaluators and subsequently, datasets in low resource languages?

While there have been efforts done in increasing datasets for low resource languages \cite{seacrowd, nusa}, not much work has been done to increase representation of evaluators. There are existing tools that crowdsource data annotation and evaluation like Amazon's Mechanical Turk, but they function more as outsourcing tools and do not increase the representation of evaluators in the wider research community.

This paper proposes a system where developers of MT systems can connect with prospective native speakers and properly annotate and evaluate the output of the MT systems. This connection opens up long term opportunities between the developer and native speaker for collaboration on other research projects. Gamification techniques are employed in the evaluation feature to increase user engagement. The annotated output is then open sourced to the public for whatever use they deem necessary. In this system, the native speakers have more negotiating power as they can set their compensation requirements whether it be monetary, acknowledgement, or authorship in a research paper.

\section{BACKGROUND}

Large language models (LLM) is a technology that has gained popularity in recent years. Its real-world applications range from conversational agents (ChatGPT), intelligent tutoring systems (Duolingo), coding assistants (Github CoPilot), and machine translation systems. Due to the wide applications of this technology, it is imperative that their developers conduct proper evaluation methods to ensure correctness, safety, and fairness for the users involved. Evaluating LLMs depends largely on the context it is being used. If the metrics being used to evaluate can be calculated, automated evaluation can be done, otherwise some manual human evaluation must be incorporated \cite{surveyLLMEval}.

\subsection{Automated Evaluation}
In an automated evaluation scenario, metrics can be categorized in to string-based or machine learning-based metrics. String-based metrics measure the distance of the characters between the reference translation and target sentence. Popular string-based metrics include BLEU, NIST, METEOR, and chrF \cite{machinetranslate_metrics}. These metrics have been typically used in evaluating MT systems developed for the annual Workshops on Machine Translations \cite{indic-findings, spanish-findings}. Machine learning-based metrics use sentence embeddings and calculates the difference between the generated text and the reference translation \cite{machinetranslate_metrics}. These metrics typically require a model trained on data from both the source and target languages. Popular machine-learning based metrics include COMET which was designed for evaluating multilingual MT systems \cite{comet}, YiSi, and BERTscore, which both tends to correlate better with human evaluation \cite{bert, yisi}. All of these metrics can provide a surface-level view of the translation without taking into account the specific language’s semantics. As such, these automated metrics may have poor correlation with human evaluation in certain scenarios. For example, the BLEU metric is a score that designates the quality of a translation. The score is calculated by measuring the precision of unigrams, bigrams, trigrams, and 4-grams of the generated translation to the high quality reference translation, and then computes a penalty score for sentences that are too short \cite{bleu}.

For example, given a source sentence in Filipino of "Ang ganda ng bahay na ito.", an MT generated english translation of: "A beautiful house this is", and a high quality reference translation of: "This is a beautiful house". The resulting BLEU score would reflect a high quality translation even though from a human's perspective, the MT output is not semantically sound.

\subsection{Human/Manual Evaluation}

Manual Evaluation involves incorporating techniques where humans intervene in the evaluation of the output \cite{auto-manual-eval}. Human judgment is considered to be one of the most reliable criterion of translations generated for human use because the real world comprehension allows the judges to give an accurate estimate of the importance of translation errors, and in turn provide adequate feedback \cite{sanders2011}. Several criteria must be considered in selecting a human evaluator such as if they are monolingual or bilingual, and their familiarity of the source texts \cite{auto-manual-eval}. For an evaluation to be effective, experienced translators are preferred \cite{laubli2018}, and there should be reasonable text volume and uninterrupted task performance \cite{przybocki2011}.

Manual evaluation is also categorized into two categories, directly expressed judgment (DEJ) and non-DEJ based. DEJ-based methods are the most common, and typical tools used include a five-point scale \cite{Callison-Burch2007}, seven-point scale \cite{Przybocki2009}, or a 100-point scale \cite{Bojar2016} to evaluate adequacy, which measures how semantically and pragmatically equivalent between the source is to the target text \cite{auto-manual-eval}, and fluency, which pertains to the text's grammaticality and naturalness \cite{auto-manual-eval}. Higher-valued scales are typically preferred for finer grained statistical analyses. In all cases, it must be clear to the evaluator what each value corresponds to in terms of quality in the scale. Non DEJ-based evaluation techniques involve using semiautomated metrics, or by humans analyzing and correcting MT outputs \cite{auto-manual-eval} such as Postediting. 

Postediting can be classified into Full, and Light. Full postediting is where the translation is completely rewritten to be stylistically normal with correct grammar and punctuation \cite{Massardo2016}, while Light postediting uses only necessary changes to make the output more comprehensible but not necessarily stylistically compelling \cite{Massardo2016}. If a gold standard human translation is not present, the postediting procedure consists of detecting translation errors between the source and target text, detecting linguistic errors in the target text, and rewriting and proofreading \cite{auto-manual-eval}. The following guidelines are suggested for effective postediting \cite{Massardo2016}:

\begin{enumerate}
        \item Aim for a semantically correct translation
        \item No information must be added or ommitted
        \item Remove or Rewrite inappropriate content
        \item Use as much of the raw MT output as possible
        \item Ensure correct spelling
        \item Omit stylistic corrections
        \item Do not restructure sentences to solely improve the natural flow of the text
\end{enumerate}

Disadvantages of manual evaluation include subjectiveness, cost, reproducibility, and low interannotator agreement (the level at which annotators are consistent with their previous judgments). Crowdsourcing direct assessment addresses these and ensures reliability of the evaluations through quality control items. The items serve as intervention between the MT outputs under evaluation, and the good and bad reference translations \cite{Graham2015}.

\subsection{The Lack of Human Representation in Low Resource Languages}

Since the majority of NLP research is centered around the English language \cite{sogard}, it's reasonable to believe that there would be an abundance of English-speaking human evaluators. Conversely, since there are lacking datasets in LRLs, human evaluators for those languages would be difficult to come by as well. An instance of this occurred in a recent submission of the Shared task in Spanish languages for the Workshop on Machine Translation in 2024 \cite{spanish-wmt}. The authors conceded that while automated metrics like BLEU provided some key information about the accuracy of their MT system for Spanish, Aragonese, Asturian, and Aranese, they could not verify if the translations are syntactically or semantically correct since none of them are able to speak these languages. \cite{spanish-wmt}.

Data scarcity is also downstream from the lack of human representation in LRLs. Few human speakers for an LRL would also mean few annotated datasets and linguistic resources needed to train MT systems \cite{indic-findings}. As an example, in the Manipuri language of Northeastern India, this data scarcity leads to difficulties in developing systems for Machine Translation and other NLP applications \cite{indic-findings}.

There are many existing crowdsourcing platforms to recruit human participants for data annotation like Amazon's Mechanical Turk (MTurk), and Clickworker. MTurk is a crowdsourcing marketplace to virtually outsource jobs to a distributed workforce \cite{mturk}.  Clickworker is a similar platform but offers a rigorous qualification check for the annotators such as tests and collection of relevant personal data \cite{clickworker}. These platforms offer a good solution for researchers seeking to recruit annotators for their data. Annotators are paid only in a certain monetary amount depending on the size and type of the job, and are rarely in direct communication with the researchers and developers, because of this ,annotators for these platforms act as outsourced labor rather than potential long term collaborators.

\subsection{Data Scarcity in Low Resource Languages}

Low Resource Languages (LRL) are languages that are typically lacking in representation in research and quality datasets. In Southeast Asia (SEA), a region with over 600 million people and 1000 indigenous languages, significant lack of resources in text, images, and other data cause poor performance of AI models \cite{seacrowd}. One of the more common and effective approaches to retrieve training data has been through online document scraping and translation \cite{nusax, indobert}. However, a recent case study on Indonesian languages showed limitations of this approach \cite{nusa}. Analysis of Wikipedia as a data source showed multiple issues such as lack of lexical diversity, a significant presence of boilerplates, and that it only contained a small subset of Indonesian languages, despite its large amounts of data. 

A different approach of generating datasets through sentence translation by native speakers was explored. The annotators translate the source language data to 11 local Indonesian languages while maintaining the sentence's sentiments, named entities, and completeness of the text. The result is a corpora of 72,444 sentences with adequate representation of multiple Indonesian languages. The dataset was also shown to be rich in lexical diversity and is more in line with colloquial Indonesian writing.\cite{nusa}.

\subsection{Gamification Principles}

Gamification is a process of introducing game elements such as scoring, ranking, and badges, in non-gaming scenarios with the goal of improving user experience and engagement. These game elements serve as motivational drivers of human behavior which can be both positive and negative. When done effectively, gamification can reduce the required cognitive resources for a certain activity by leveraging the reward and emotional response of the individual \cite{gamification}. The goal setting theory provides a conceptual framework for mapping certain principles to gamification \cite{goals}. For the sake of this paper, only the following principles from the said framework are explored:

\begin{enumerate}
        \item Specific goals – To focus the user’s attention, goals must be clear in how to achieve them. These could be done through badges, leaderboards, progress bars, or levels.
        \item Self-efficacy – The system must provide feedback mechanisms or purposeful elements that allows the user to feel responsible for their success and provide a larger context for their achievement.
        \item Ability – Complex tasks must be divided into smaller tasks and are achievable at the user’s current level. The system must provide mechanisms to assist the user to complete the task should they get stuck.
        \item Goal commitment – The system must show the importance of the task and feel committed to it. Popular game elements include narrative, story, social network and collaboration.
\end{enumerate}

\section{PLATFORM FEATURES}

\subsection{World Map Visualization for Datasets and Evaluators}

\begin{figure}[h]
  \centering
  \includegraphics[width=\linewidth]{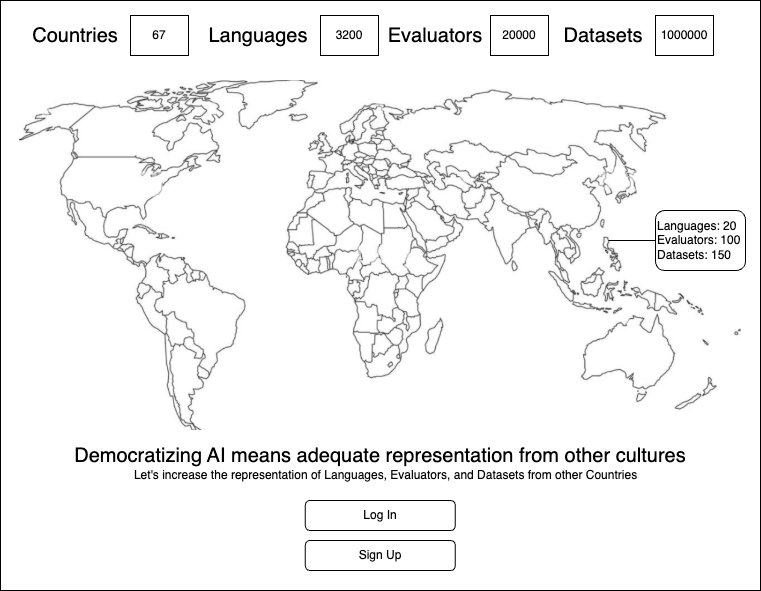}
  \caption{The landing page of the platform serves as an interactive dashboard of each country's evaluators, languages, and datasets}
  \Description{landing page showing visualization map}
  \label{fig:registration}
\end{figure}

\begin{figure}[h]
  \centering
  \includegraphics[width=\linewidth]{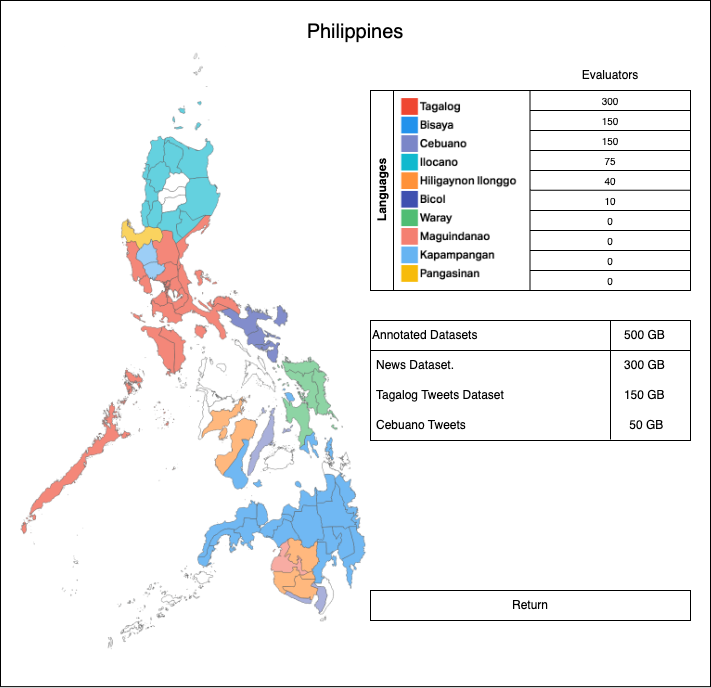}
  \caption{The zoomed in view of the Country's data from the landing page}
  \Description{country landing page showing visualization map}
  \label{fig:drilldowncountry}
\end{figure}

The landing page of the platform as shown in Figure \ref{fig:registration} is an interactive world map which serves as a data visualization dashboard of the representation of countries, languages, evaluators, and datasets that the platform currently supports. The user can hover over any country and it shows the number of evaluators, languages, and annotated datasets belonging to that country. Clicking on a specific country shows a dashboard showing the number of languages supported by the platform, current annotated datasets available, and the number of human evaluators available (see Figure \ref{fig:drilldowncountry}).  To maintain privacy, the names of the evaluators are not shown. This feature lays out the specific goals of this platform which is to increase representation of evaluators and in effect, annotated data for underrepresented languages from other countries. 

\begin{figure}[h]
  \centering
  \includegraphics[width=\linewidth]{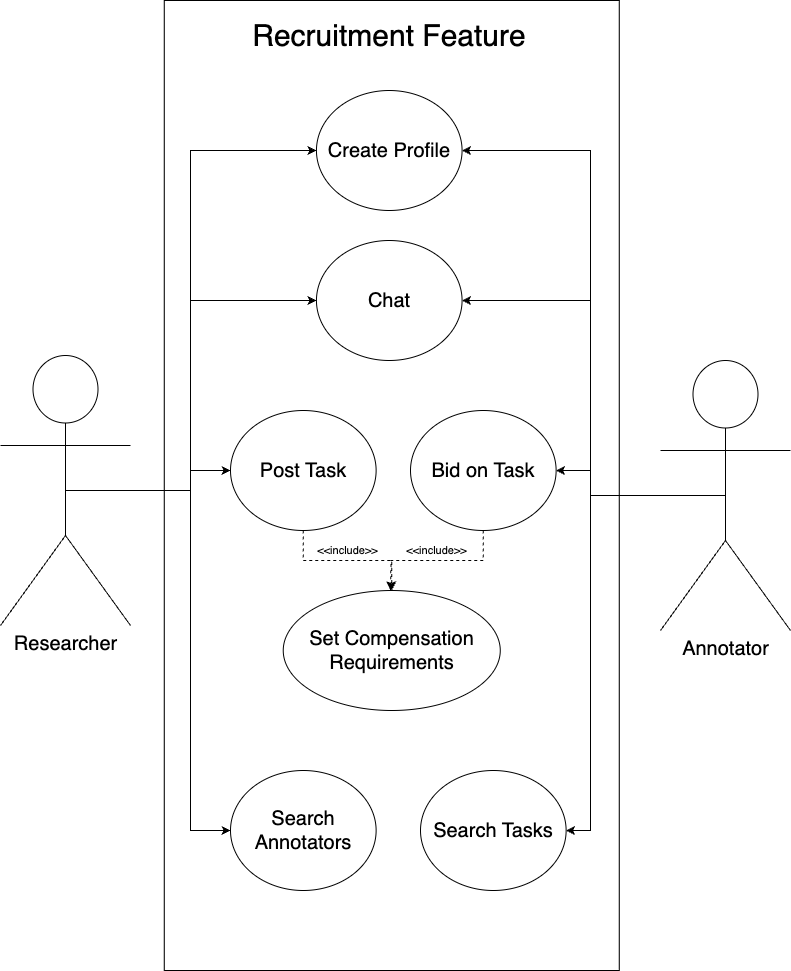}
  \caption{A UML Use Case Diagram of the Recruitment Feature showing the tasks available to the Researcher and Annotator User}
  \Description{UML recruitment feature}
  \label{fig:recruitment}
\end{figure}

\subsection{Recruitment Feature}

One of the key aspects of the platform is to provide researchers with an avenue to connect with linguistic experts or native speakers of a low resource language for them to perform certain evaluation tasks. The recruitment feature will have to user types: A “Researcher” developing an MT system, and an “Annotator” who will evaluate the output of the MT system. A UML use case diagram for this feature is shown in Figure \ref{fig:recruitment}. The functional requirements of this feature are the following:

\begin{enumerate}
        \item The system shall provide the Researcher and Annotator the ability create their Profiles and list the languages they are interested in working on.
        \item The system shall provide the Researcher the ability search for individuals with the necessary language expertise (Annotator), and vice versa
        \item The system shall provide the Annotator the ability to search for projects in need of Human Evaluators
        \item The system shall provide the Researcher the ability to request, accept, or deny a connection  request to the Annotator and vice versa
        \item The system shall provide a chat feature between the Researcher and Annotator once the connection request is accepted.
        \item The system shall provide the Annotator and Researcher the ability to set their compensation requirements/benefits when posting or bidding on a task (monetary, paper authorship, etc.)
        
\end{enumerate}

To maintain privacy of the users' data when searching for Researchers and Annotators, only the username, language interested, and certificates are shown. Certificates will only be listed in a generic manner to not show any personal information (i.e. "Earned Masters in Filipino Studies"). Only when the connection between the two is accepted can they choose to share these details with one another.

\subsection{Evaluation Feature}

\begin{figure}[h]
  \centering
  \includegraphics[width=\linewidth]{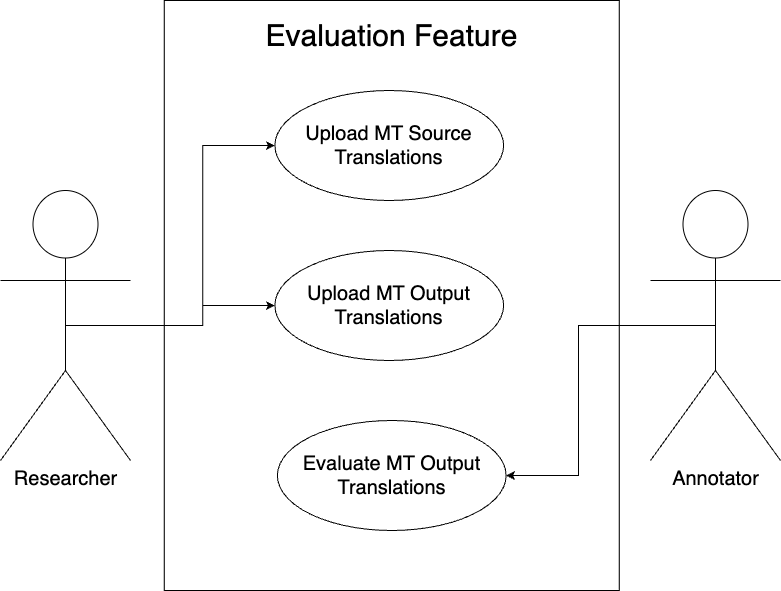}
  \caption{A UML Use Case Diagram of the Evaluation Feature showing the tasks available to the Researcher and Annotator User}
  \Description{UML use case for evaluation}
  \label{fig:evaluation}
\end{figure}

For the Evaluation feature, the system will primarily focus on a DEJ-based evaluation of Adequacy and Fluency of the generated translation as tasks that the Annotator will perform. Each of these tasks will incorporate gamification design strategies for enhanced user experience. The UML use case diagram is shown in Figure \ref{fig:evaluation}.

\begin{enumerate}
        \item The platform shall provide an upload feature that allows the Researcher to provide all the source sentences and their corresponding generated translations, and specify which language or country it belongs to.
        \item The platform shall provide a view of each of the generated translations and the source sentences to the Annotator.
        \item The Annotator uses a slider to rate from 1 (lowest) - 100 (highest) the translation for its Adequacy, Fluency
        \item The platform shall allow the Annotator to rewrite or postedit the translation to provide a more adequate representation of the original sentence
        \item The platform shall provide an information tooltip to describe the metrics of Adequacy and Fluency, as well as guidelines on how to do a Rewrite (see Section 1.2).
        \item The platform shall provide the results of all the evaluations by the Annotator to the Researcher.
        \item Once evaluation is completed, the platform shall open source the evaluated data, allowing all users of the platform to use the data for whatever purpose.
\end{enumerate}

\begin{figure}[h]
  \centering
  \includegraphics[width=\linewidth]{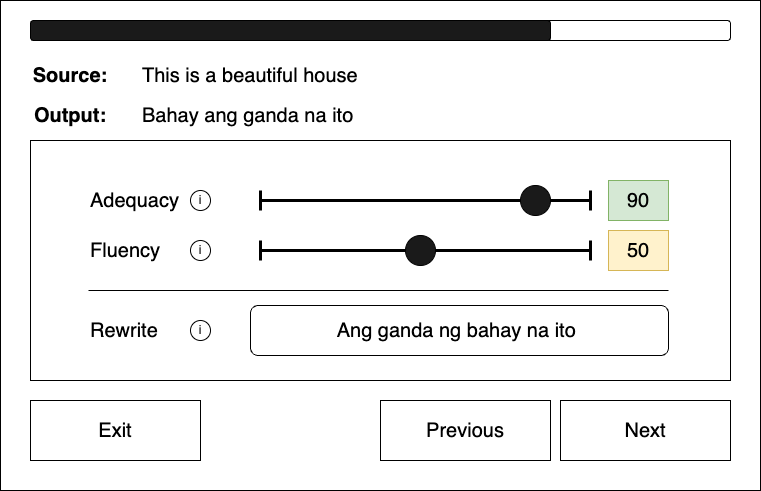}
  \caption{A wireframe for the evaluation of the MT output}
  \Description{wireframe figure}
  \label{fig:wireframe}
\end{figure}

Figure \ref{fig:wireframe} shows a wireframe for the evaluation feature of the MT output. The Annotator is given the source sentence and output translation, and a 100-point slider to rate for Adequacy and Fluency. Below the slider is an option for the Annotator to lightly postedit the output in to a more appropriate translation before proceeding to the next or previous translation. 

To discourage the use of AI-generated tools for postediting, the system disables the copying of the source and output text to the clipboard, and pasting of any text to the Rewrite text input. GPTZero's API for detecting AI-generated text will also be integrated. The system will show an error if the API determines the postedit to be AI-generated text. 

A progress bar is also shown above to give the user feedback on how much has been accomplished already for the Researcher's task. This feature enforces the "Ability" gamification principle because it divides the large task of annotating all of the MT outputs in to tasks that are completed one at a time with the option of pausing in the middle should the user want to take a break.

\begin{figure}[h]
  \centering
  \includegraphics[width=\linewidth,height=7cm,keepaspectratio]{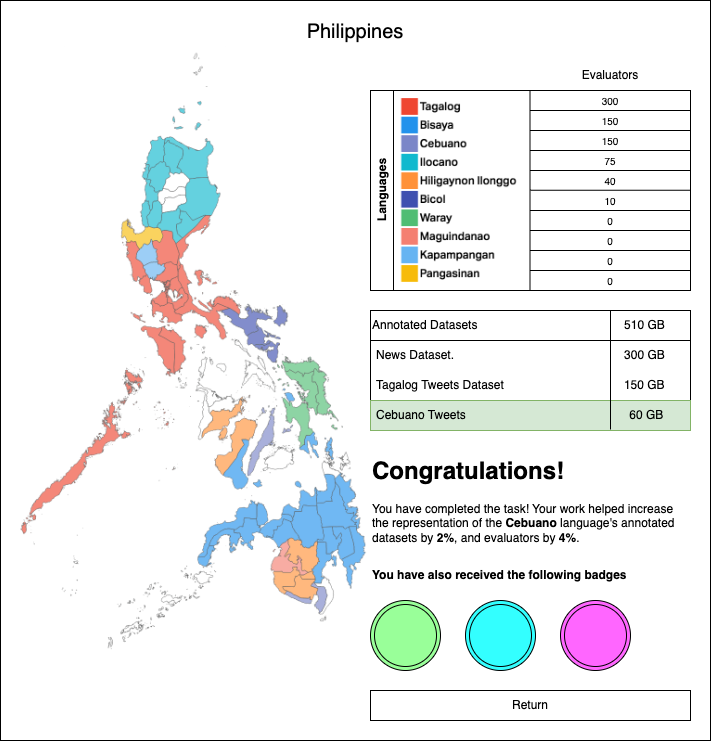}
  \caption{The results page after the Annotator completes the task}
  \Description{results page}
  \label{fig:results}
\end{figure}

Once accomplished, the results page shown in Figure \ref{fig:results} is where the system presents the summary of the Annotator's actions, and the positive effects of the task completion on the representation of datasets and evaluators on the specific language. Certain badges may also be rewarded to the Annotator depending on the task. This results page enforces the gamification principles of "Self efficacy" and "Goal commitment". It shows the importance of the task and provides feedback on how their achievement makes a positive impact on the representation of evaluators and data.

\subsection{Badges and Leaderboard}

An important aspect of gamification is the addition of Badges earned, and a Leaderboard feature. For the sake of this paper, no specific badges were discussed but a proposed design of the badge system would be that the more badges earned, the higher placement of the annotator on the leaderboard. A high placement on the leaderboard and would result in a higher placement on the search results for annotators. Badges of higher value would come from tasks from languages with fewer datasets and evaluators, encouraging the accommodation of and participation in more lower resource languages.

\section{DISCUSSION}

Lack of data representation is a longstanding issue for low resource languages. Much research efforts have been done to address this, but an equally important issue to consider must be increasing representation of human evaluators for these languages. Before we can develop novel human-centered methods in evaluation, there has to be humans to use those methods. Human evaluators, especially those with the necessary expertise, can contribute the necessary knowledge of the unique syntax and semantics of the language to guide the design of MT systems, and LLMs in general. This design proposal serves as a possible tool to connect with these individuals or serve as and area of interest for researchers to look in to for designing similar platforms.

\subsection{Evaluating this Platform}

\begin{figure}[h]
  \centering
  \includegraphics[width=\linewidth,height=7cm,keepaspectratio]{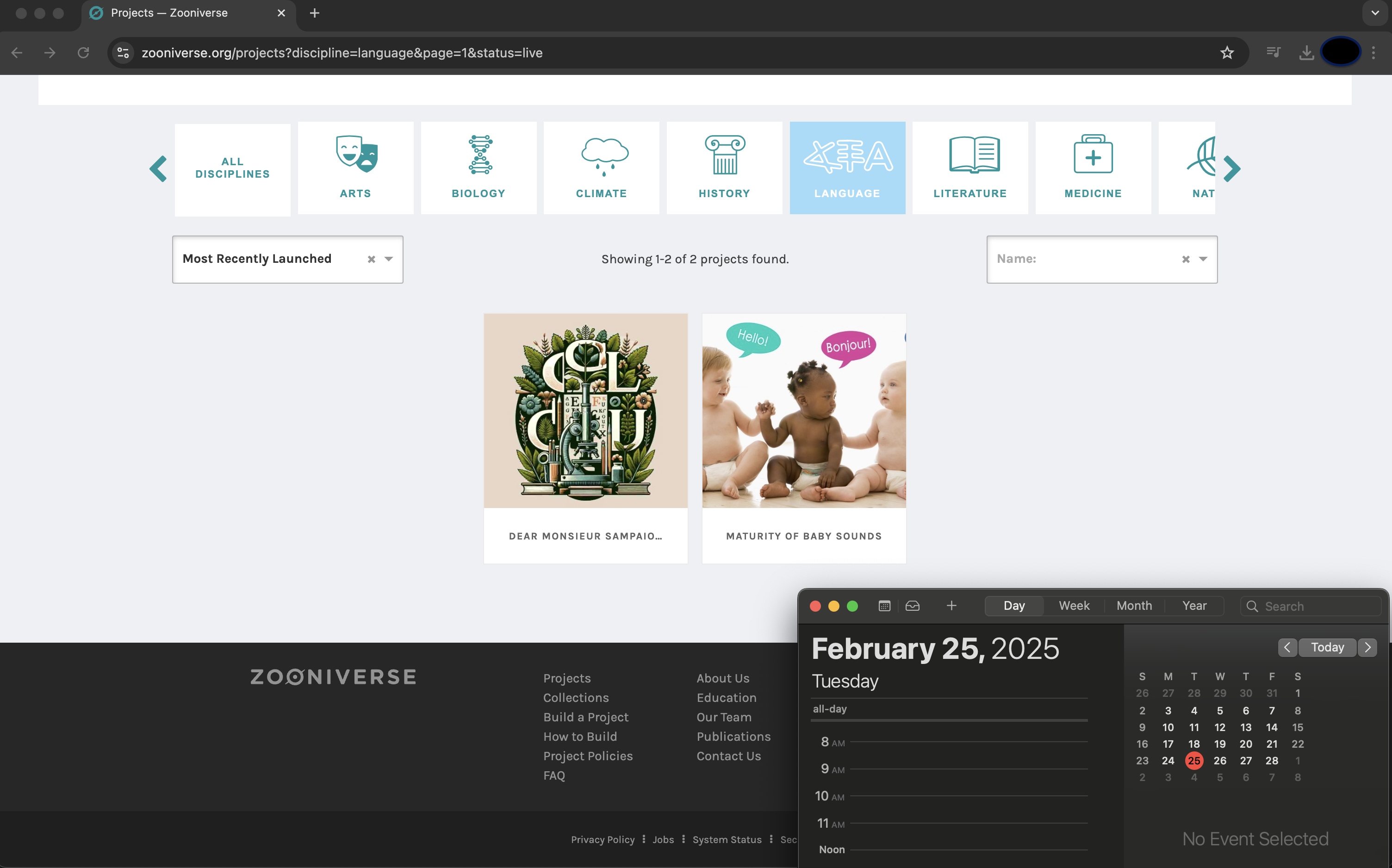}
  \caption{Screengrab of the Zooniverse Platform's Language Projects as of Feb 2025}
  \Description{results page}
  \label{fig:zooniverse}
\end{figure}

A proposed metric for the success of this platform would be its Daily Active Users (DAU), Session Duration (the time users spend on the platform), User Acquisition Rate (number of new users over a specific period of time), Conversion Rate (percentage of users completing a desired action), and number of tasks posted compared to other similar recruitment platforms such as Freelancer, Upwork. The closest similar platform would be Zooniverse which is catered to recruitment for research work. Its research topics include Biology, Literature, Physics, and Social Science, among others. However for Natural Language Processing research, it’s lacking in such a way that as of February 2025, there are only 2 available projects for language research as seen in Figure \ref{fig:zooniverse}.

To be able to achieve a high DAU, Session Duration, User Acquisition Rate, and Conversion Rate, the developers of this platform must be in constant communication with its users, the Researchers and Annotators, and receive feedback from them to guide their development.

An effective marketing campaign is also needed building an adequate user base for the platform. Since the goal of the platform is to increase representation in low resource languages, there needs to be a coordinated effort in reaching out to these individuals. A possible plan of action would be to market to universities' linguistics or natural language processing (NLP) departments or NLP conferences in regions where there are an abundance of LRLs like AACL (Asia-Pacific Chapter of ACL), AfricaNLP, and other Low-Resource NLP Workshops like WMT (Workshops on Machine Translation).

\subsection{Challenges}

An anticipated challenge in this type of platform would be the verification of the expertise of the Annotators. The quality of the would-be open sourced data is heavily dependent on the quality of the annotators. While there are many existing methods for verifying a user's identity, there isn't a universal method for verifying a user's language expertise. For certain languages, there are standardized certifications like JLPT (Japanese), TOPIK (Korean), and TOEFL/IELTS (English) which can serve as a signal of expertise, but for the majority of languages, especially low resource ones, none exist. An option for these scenarios in the meantime would be for formal education certificates or diplomas (Bachelors/Masters Diploma in English, Filipino Studies, African Studies, etc) to serve as expertise verifications.

\subsection{Future Work}

To attempt to launch this platform with the goal of getting a global user base would be unfeasible. A pilot study in a smaller region would be a more sensible approach. For example, in the Philippines, Cebuano is the lingua franca of the Visayas region, and is considered to be a much lower resource language compared to Tagalog which is typically centered in the Luzon region. The study would involve marketing to both Tagalog and Cebuano speakers, with the latter being heavier, and comparing the user engagement and annotated output for both. The results of this study would guide future studies and development surrounding this platform.

Another future work to consider would be the integration of other evaluation tasks for other NLP tasks, and their subsequent gamification. Bias mitigation in text and story generation is an important task for humans to evaluate. The annotator would be given the prompt and the generated text to search for cultural stereotypes and misrepresentations. Evaluating datasets used for training would also be an important next step as it is known to heavily influence the output of any MT system. A random sample of the training data would be presented to the Annotators for them to analyze for over representation of English words, or cultural stereotypes.

While postediting is a useful tool for MT systems evaluation, and is already an established practice in the translation pipeline in many professional contexts \cite{mt-effort}, there still remains a lot of research questions surrounding it such as in determining the cognitive effort needed to perform such tasks \cite{Lacruz2014}, and determining the feasibility of postediting being used in different language pairs \cite{mt-effort}. This tool, with its postediting feature can assist other researchers in exploring this question.

\bibliographystyle{ACM-Reference-Format}
\bibliography{sample-base}

\end{document}